\documentclass[final,5p,times,twocolumn]{elsarticle}

\usepackage{amssymb}
\usepackage{amsmath}
\usepackage{multirow}

\journal{Neurocomputing}

\begin{document}

\begin{frontmatter}



\title{HRPVT: High-Resolution Pyramid Vision Transformer for medium and small-scale human pose estimation}




\author[1]{Zhoujie Xu}
\ead{236142147@mail.sit.edu.cn}


\author[1]{Meng Dai \corref{cor1}}
\ead{daimeng@sit.edu.cn}

\author[1]{Qing Zhang}
\ead{zhangqing0329@gmail.com}

\author[1]{Xiaodi Jiang}
\ead{297973051@qq.com}

\affiliation[1]{organization={School of Computer Science and Information Engineering, Shanghai Institute of Technology},
            addressline={100 Haiquan Road},
            city={Shanghai},
            postcode={201418},
            country={China}}
\cortext[cor1]{Corresponding author}

\begin{abstract}

Human pose estimation on medium and small scales has long been a significant challenge in this field. Most existing methods focus on restoring high-resolution feature maps by stacking multiple costly deconvolutional layers or by continuously aggregating semantic information from low-resolution feature maps while maintaining high-resolution ones, which can lead to information redundancy. Additionally, due to quantization errors, heatmap-based methods have certain disadvantages in accurately locating keypoints of medium and small-scale human figures. In this paper, we propose HRPVT, which utilizes PVT v2 as the backbone to model long-range dependencies. Building on this, we introduce the High-Resolution Pyramid Module (HRPM), designed to generate higher quality high-resolution representations by incorporating the intrinsic inductive biases of Convolutional Neural Networks (CNNs) into the high-resolution feature maps. The integration of HRPM enhances the performance of pure transformer-based models for human pose estimation at medium and small scales. Furthermore, we replace the heatmap-based method with SimCC approach, which eliminates the need for costly upsampling layers, thereby allowing us to allocate more computational resources to HRPM. To accommodate models with varying parameter scales, we have developed two insertion strategies of HRPM, each designed to enhancing the model's ability to perceive medium and small-scale human poses from two distinct perspectives. Our proposed method achieved scores of 76.3 AP and 75.5 AP on the MS COCO Keypoint validation and test-dev datasets, respectively, while reducing the number of parameters by 60\% and the GFLOPs by 62\% compared to HRNet-W48. Furthermore, our method achieved the higher score in the AP$^{M}$ metric among most state-of-the-art methods, validating its superiority in medium and small-scale human pose estimation.

\end{abstract}


\begin{highlights}
\item A novel hybrid Vision Transformer architecture has the capability to enhance human pose estimation at medium and small scales.
\item Two insertion strategies derived from HRPM enhance the estimation of human poses at medium and small scales from two different perspectives.
\item HRPVT achieves superior performance while reducing the number of parameters by 60\% and the GFLOPs by 63\% compared to HRNet-W48.
\end{highlights}


\end{frontmatter}




\section{Introduction}
Human pose estimation (HPE) stands as a cornerstone task in computer vision, involving the detection of keypoint locations on the human body and the categorization of these keypoints for each individual in a given image. Its significance extends to numerous downstream applications, such as activity recognition \cite{ni2017learning,baradel2018glimpse,liu2017skeleton}, human-robot interaction \cite{mazhar2018towards,zhang2020empowering}, and video surveillance \cite{hattori2018synthesizing,varadarajan2018joint}. However, HPE presents formidable challenges owing to a multitude of factors, including occlusion, truncation, under-exposed imaging, blurry appearances, and the low-resolution nature of person instances.
In the early stages of 2D human pose estimation, regression-based methods were frequently explored. These methods directly regress the keypoint coordinates within a computationally efficient framework. However, due to unsatisfactory performance, only a limited number of existing methods have adopted this scheme. In recent years, heatmap-based methods have emerged as the predominant approach, offering advantages such as reducing false positives and facilitating smoother training through the assignment of probability values to each position. Despite their success, these methods encounter a significant challenge: the persistent quantization error problem, particularly evident in scenarios with low resolution or small-scale persons. This issue arises from the mapping of continuous coordinate values into discretized 2D downscaled heatmaps. Numerous efforts have been made to address this quantization error, with Simcc \cite{li2022simcc} being one notable example. Simcc approaches human pose estimation as two classification tasks for horizontal and vertical coordinates, which not only removes costly upsampling layers but also demonstrates superior localization accuracy for medium and small-scale persons compared to heatmap-based methods.

Recently, Vision Transformer (ViT) \cite{dosovitskiy2020image} has revolutionized the use of pure transformer architecture in vision tasks, showcasing promising outcomes. While ViT excels in tasks like image classification, directly adapting it to pixel-level dense predictions, such as human pose estimation, poses significant challenges. For example, ViTPose \cite{xu2022vitpose} achieves superior performance but at the cost of substantial model parameter size and computational complexity. Addressing this, the PVT series \cite{wang2021pyramid,wang2022pvt} introduces pioneering methods, including a progressive shrinking pyramid and spatial-reduction attention to refine the pure transformer architecture and improve its suitability for dense prediction tasks. Although the PVT series significantly enhances the pure transformer architecture, it still lacks the intrinsic inductive biases of CNNs, which excel in modeling local visual structures and handling scale variance. As a pure transformer, it boasts strong capabilities in modeling long-range dependencies but falls short in capturing local feature details inherent to CNNs. For HPE, on one hand, keypoints constraint relationships and visual cues precisely correspond to these two critical abilities \cite{li2021tokenpose}; on the other hand, in certain scenarios, such as outdoor extreme sports, the protean nature of human poses makes it difficult to accurately predict medium and small-scale human postures, even when using a top-down paradigm. Moreover, in many previous works, such as Higher-HRNet \cite{cheng2020higherhrnet}, high-resolution representations with high quality have been proven to be significantly helpful in addressing this issue. Thus, we wondered if it is possible to eliminate costly upsampling layers through SimCC and instead focus our limited computational resources on integrating the intrinsic inductive biases of CNNs, such as scale-invariance and locality, into the high-resolution feature maps of PVT. This approach may achieve a favorable trade-off between accuracy and efficiency.

Motivated by this, we propose a novel model called the High-Resolution Pyramid Vision Transformer (HRPVT), which uses a combination of PVT v2 and Simcc as the baseline. Building on this foundation, we have designed a module, namely the High-Resolution Pyramid Module (HRPM). The HRPM consists of two sub-modules, HRPM v1 and HRPM v2, which are designed to model the scale-invariance and locality by CNNs in the high-resolution feature maps, thereby enhancing the network's ability to localize keypoints in medium and small-scale human figures. Moreover, according to the insertion position of HRPM v2, we designed two insertion strategies, namely Layer-wise Insertion and Stage-wise Insertion, to accommodate baselines with different capacity from two perspectives.Our main contributions are summarized as follows:
\begin{itemize}
    \item We propose a novel model called HRPVT, which introduces the intrinsic inductive biases of scale invariance and locality, inherent in CNNs, into the high-resolution feature maps of PVT v2. This enhances the network's ability to localize keypoints in medium and small-scale human figures.
    \item We develop two insertion strategies—Layer-wise Insertion and Stage-wise Insertion—each designed from a distinct perspective to address baselines of varying complexity.
    \item Our HRPVT model has demonstrated outstanding performance on both the MS COCO and MPII datasets. Notably, on the MS COCO dataset, HRPVT outperformed HRNet-W48 while using only 40\% of its parameter count and 37\% of its GFLOPS, achieving even better results.
\end{itemize}

\section{Related work}
Methods for 2D Human Pose Estimation (HPE) focus on determining the 2D coordinates or spatial positioning of human body keypoints in images or videos. Two main deep learning strategies are utilized: regression and heatmap-based approaches.
\subsection{Regression-based methods}
Regression methods employ a comprehensive framework that learns to directly map the input image to the locations of body joints or to parameters defining human body models. One of the trailblazing works in this domain is DeepPose \cite{toshev2014deeppose}, which pioneered the transformation of the human pose estimation challenge into a keypoint coordinate regression problem, sparking a series of influential subsequent studies e.g., \cite{carreira2016human,li2021pose,fan2015combining,luvizon2019human,luvizon20182d}. However, ongoing research uncovered several issues. First, the extensive numerical range and dispersed distribution of human keypoint coordinates make direct learning by the network difficult. Second, there is a wealth of constraint information both among human keypoints and between humans and their environment, but this information is lost in coordinate regression methods. These deficiencies substantially hinder the effectiveness of coordinate regression techniques and prevented them from outperforming heatmap-based methods. This situation persisted until Li et al. \cite{li2021human} ventured into probabilistic modeling by proposing a normalizing flow model named RLE (Residual Log-likelihood Estimation). This model is designed to capture the distribution of joint locations and seeks optimized parameters through residual log-likelihood estimation.
\begin{figure*}[t]
\centering
\includegraphics[width=\linewidth]{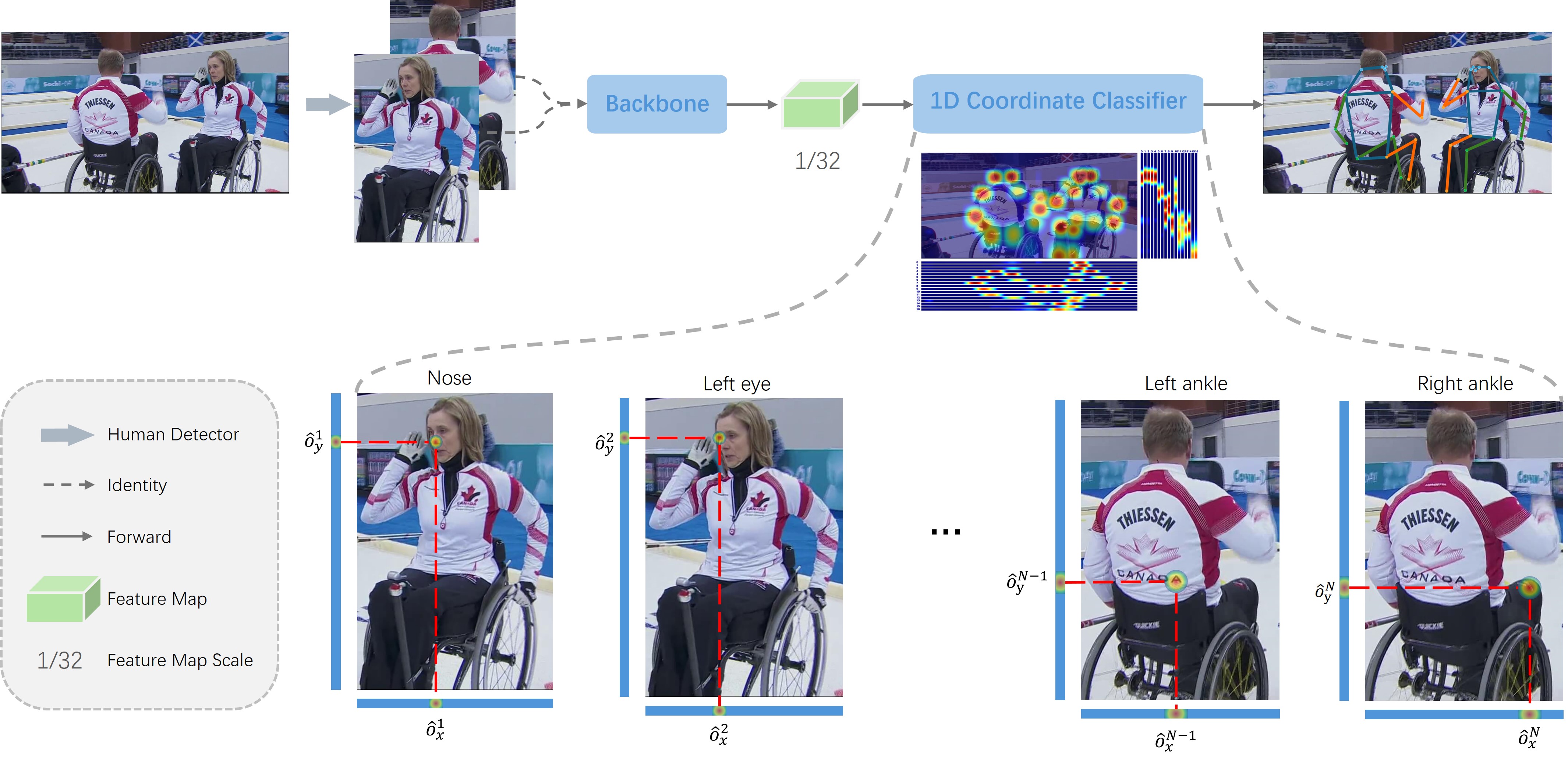}
\caption{The pipeline of HRPVT. Given an input image, a human detector is first applied to generate a set of human bounding boxes. Subsequently, the backbone of HRPVT is used to extract keypoint representations of the human body. Finally, the 1D Coordinate Classifier, SimCC, predicts the detailed localization of keypoints for each individual.
}\label{fig: overall}
\end{figure*}
\subsection{Heatmap-based methods}
Heatmap-based approaches in human pose estimation (HPE) diverge from directly pinpointing the 2D coordinates of human joints. Instead, these methods focus on generating 2D heatmaps, which are formulated by superimposing 2D Gaussian kernels over each joint's location. This strategy not only retains precise spatial information about each joint but also facilitates a smoother training process, offering a distinctive advantage over methods that aim to estimate joint coordinates directly. The use of heatmaps for representing joint locations has seen a surge in interest, prompting the development of CNNs’ architectures tailored for this purpose. Wei et al. \cite{wei2016convolutional} introduced the Convolutional Pose Machines, a multi-stage framework that excels in pinpointing keypoint locations by leveraging 2D belief maps from prior stages to enhance prediction accuracy. Concurrently, Newell et al. \cite{10.1007/978-3-319-46484-8_29} developed the stacked hourglass network, a creative encoder-decoder architecture that iteratively captures and processes body pose information, supplemented by intermediate supervision. Sun et al. \cite{sun2019deep} contributed significantly with the High-Resolution Network (HRNet), which maintains high-resolution representations throughout the network by connecting multiresolution subnetworks and performing multi-scale fusions. This innovation greatly enhances the accuracy of keypoint heatmap predictions. Although the heatmap-based approaches are widely used, quantization error still remains a significant challenge, especially with low-resolution inputs. Cheng et al. \cite{cheng2020higherhrnet} introduced an enhancement to HRNet, termed Higher Resolution Network, which employs deconvolution on the high-resolution heatmaps produced by HRNet to relatively reduce the quantization error and significantly boosts the detection of medium- and small-scale individuals. 

\subsection{Quantization error problem}
To mitigate the significant quantization error arising from the discretized 2D downscaled heatmaps. Zhang et al. \cite{zhang2020distribution}  suggested using a Taylor-expansion based distribution approximation for post-processing. This method effectively incorporates the distribution information of heatmap activations. Gu et al. \cite{gu2021removing} proposed a method for obtaining coordinate values by normalizing the feature map and then computing the expectations. Li et al. \cite{li2022simcc} presented a novel approach by reformulating HPE into two distinct classification tasks, one for horizontal coordinates and the other for vertical. They introduced the SimCC method, which significantly improves sub-pixel localization accuracy and minimizes quantization errors through the uniform segmentation of each pixel into several bins.
\subsection{Vision Transformers with inductive bias}
ViT \cite{dosovitskiy2020image} stands out as a groundbreaking endeavor in applying a pure transformer approach to vision tasks, yielding promising outcomes. The simultaneous developments of MViT \cite{fan2021multiscale}, PVT \cite{wang2021pyramid}, and Swin \cite{liu2021swin} incorporate multi-scale feature hierarchies into the transformer design, mirroring the spatial arrangement found in conventional convolutional architectures like ResNet-50. However, a notable limitation lies in these ViT-like methods' absence of intrinsic inductive bias in capturing local visual structures, relying instead on implicit learning from extensive data. DeiT \cite{touvron2021training} presents a method to distill knowledge from CNNs to transformers during training. Nonetheless, this approach necessitates employing a pre-existing CNN model as a teacher, thereby introducing additional computational overhead during training. Subsequent efforts have sought to imbue vision transformers with the intrinsic inductive bias of CNNs. For instance, \cite{li2021localvit,graham2021levit,wu2021cvt} adopt a strategy of stacking convolutions and attention layers sequentially, thereby establishing a serial structure conducive to modeling both locality and global dependencies. However, this sequential approach may inadvertently overlook the broader global context while focusing on local features (and vice versa). In contrast, ViTAE \cite{xu2021vitae} offers a novel approach by concurrently modeling locality and global dependencies through a parallel structure within each transformer layer. This parallel architecture not only enhances computational efficiency but also facilitates comprehensive understanding by capturing both local and global features simultaneously.

\section{Methodology}
The proposed HRPVT utilizes PVT v2 as the backbone to extract comprehensive representation information and employs SimCC as the 1D coordinate classifier for predicting keypoint coordinates using separate classifiers for horizontal and vertical dimensions. The structure of HRPVT is shown in Figure \ref{fig: overall}. We have further optimized this foundation by developing the HRPM, which is tasked with incorporating multi-scale contextual details into tokens and enhancing the modeling of local low-level semantic representations. The subsequent sections will provide detailed introductions to each component of the HRPVT.
\begin{figure*}[t]
\centering
\includegraphics[width=\linewidth]{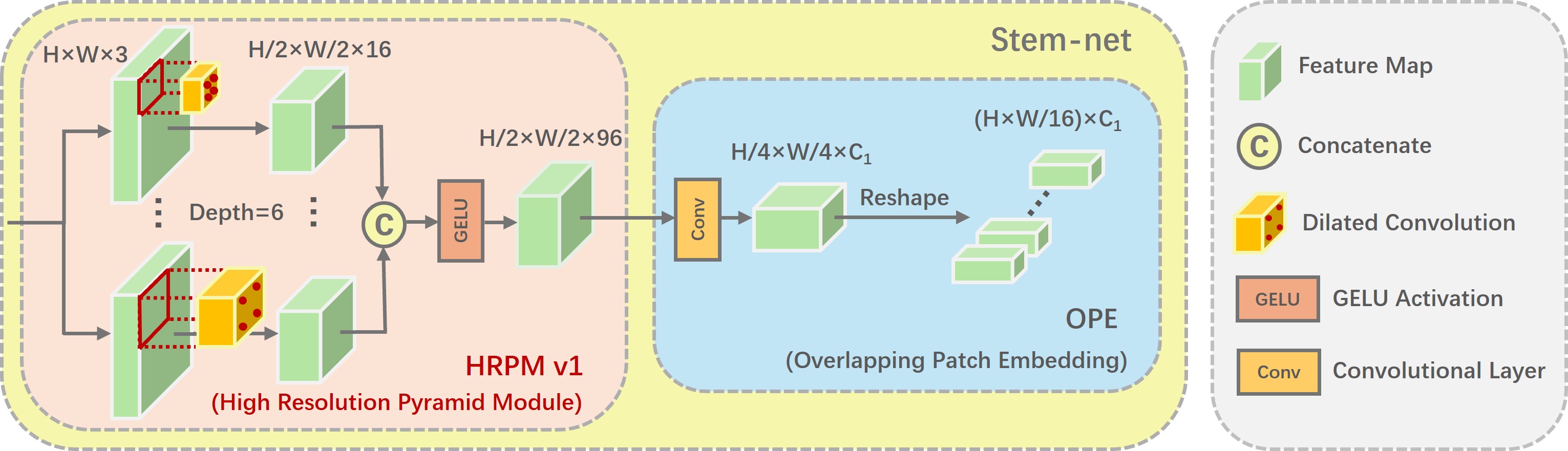}
\caption{Illustration of the structure of HRPM v1: In the stem-net, given a cropped image of size H×W×3, hierarchical hybrid-dilated convolutions are first applied to obtain six feature maps with different receptive fields. These feature maps are then stacked through a concatenation operation to form a high-resolution feature pyramid of size H/2×W/2×96. Subsequently, a strided convolutional layer is used to fuse and compress the features while downsampling the feature maps to reduce computational load in subsequent stages. Finally, the feature maps are reshaped into a token sequence to serve as input for the following stages}\label{fig: hrpmv1}
\end{figure*}

\subsection{Revisiting PVT v2 and Simcc}
We now give a brief review of our baseline, i.e. the combination of PVT v2 and SimCC. In the first stage, given an input image $x$ of size $H \times W \times 3$, overlapping patch embedding (OPE) is utilized to tokenize the images, which divides the image into $H/4 \times W/4$ patches to model the local continuity information. Subsequently, the flattened patches are fed into a linear projection, resulting in embedded token sequence of size $(H/4 \times W/4)\times C_1$, where $C_1$ represents the token dimension in the first stage, respectively. Then, the resulting tokens are fed into the following PVT v2 encoder layers. Each PVT v2 encoder layer is composed of two parts, i.e., a spatial-reduction attention (SRA) layer and a convolutional feed-forward network (CFFN). 

\paragraph{\textbf{SRA}}
In contrast to multi-head self-attention (MHSA), SRA maintains consistency with MHSA in other network architectures, with the exception that SRA reduces the spatial scale of keys and values before the attention by spatial-reduction (SR) operation. The details of $SR(\cdot)$ can be described as follows:
\begin{equation}
SR\left(s_i\right) = LN\left(Seq2Img\left(s_i, R_i\right) W^{R}\right).
\end{equation}
Here, $s_i \in \mathbb{R}^{(H_i W_i) \times C_i}$ represents the token sequence of $i^{th}$ stage, and $R_i$ denotes the reduction ratio of the attention layers in Stage $i$. $Seq2Img(\cdot)$ is an operation of reshaping the token sequence $s_i$ back to feature map of size $\frac{H_i \times W_i}{R_i^2} \times \left( R_i^2 C_i \right)$. $W^{R} \in \mathbb{R}^{(R_i^2 c_i) \times c_i}$ is a linear projection that reduces the dimension of the token sequence to $C_1$. $LN(\cdot)$ refers to layer normalization \cite{ba2016layer}.

\paragraph{\textbf{CFFN}}
In comparison with the original feed-forward network (FFN), the CFFN incorporates a 3 × 3 depth-wise convolution with a padding size of 1, enabling it to capture the local continuity of the input tensor. Additionally, the introduction of positional information through zero-padding in both the OPE and CFFN allows for the removal of fixed-size positional embeddings previously used in PVT v1.

After passing through multiple PVT v2 encoder layers, the output token sequence is transformed into a feature map $f_1$ of size $H/4 \times W/4 \times C_1$. Similarly, by utilizing the token sequence generated from the preceding stage as input, subsequent feature maps  $f_2$, $f_3$, and $f_4$ are derived, with each having strides of 8, 16, and 32 pixels respectively, relative to the original input image.

Given the feature map  $f_4$ of size $H/32 \times W/32 \times N$ extracted by PVT v2, where $N$ depends on the number of keypoints in the dataset, SimCC, which serves as the pose estimation head, first flatten $f_4$ into embeddings $e \in \mathbb{R}^{N \times (H/32 \times W/32)}$. Then, two linear projections are performed independently for the vertical and horizontal axes to encode the coordinate information of each keypoint for $N$ keypoints. The formula is as follows: 
\begin{equation}
X_i, Y_i = FC_x(e_i), FC_y(e_i).
\end{equation}
Here, $e_i \in \mathbb{R}^{H/32 \times W/32}$ stands for the embedding of the $i^{th}$ keypoint, $FC(\cdot)$ represents a fully connected layer, and $X_i \in \mathbb{R}^{W \times K}$, $Y_i \in \mathbb{R}^{H \times K}$ respectively represent encoded Simcc labels for the horizontal and vertical axes, where $K$ is the scaling factor. It should be noted that Gaussian label smoothing is used, with the standard deviation set to 6.0 by default, such that $X_i$ and $Y_i$ follow a Gaussian distribution. Subsequently, the two generated sequences $X_i$, $Y_i$ are fed into a coordinate classifier to decode the horizontal and vertical coordinate information. Specifically, the decoding process is as follows:
\begin{equation}
\hat{o}_x^i = \frac{\displaystyle \arg\max_{x_j} \, p_x^i (x_j)}{K},
\end{equation}
\begin{equation}
\hat{o}_y^i = \frac{\displaystyle \arg\max_{y_j} \, p_x^i (y_j)}{K}.
\end{equation}
Here, $x_j \in [1, X_i]$, $y_j \in [1, Y_i]$ denotes the $j^{th}$ classification bin on $X_i$, $Y_i$, respectively. $p^{i}(\cdot)$ indicates the predicted probability of the horizontal or vertical coordinate of the $i^{th}$ keypoint. $\hat{o}^i$ represents the coordinate prediction of $i^{th}$ keypoint. Finally, combining the $(\hat{o}_x^i, \hat{o}_y^i)$ pairs of $N$ keypoints results in the predicted coordinates for all keypoints.
\begin{figure*}[t]
\centering
\includegraphics[width=\linewidth]{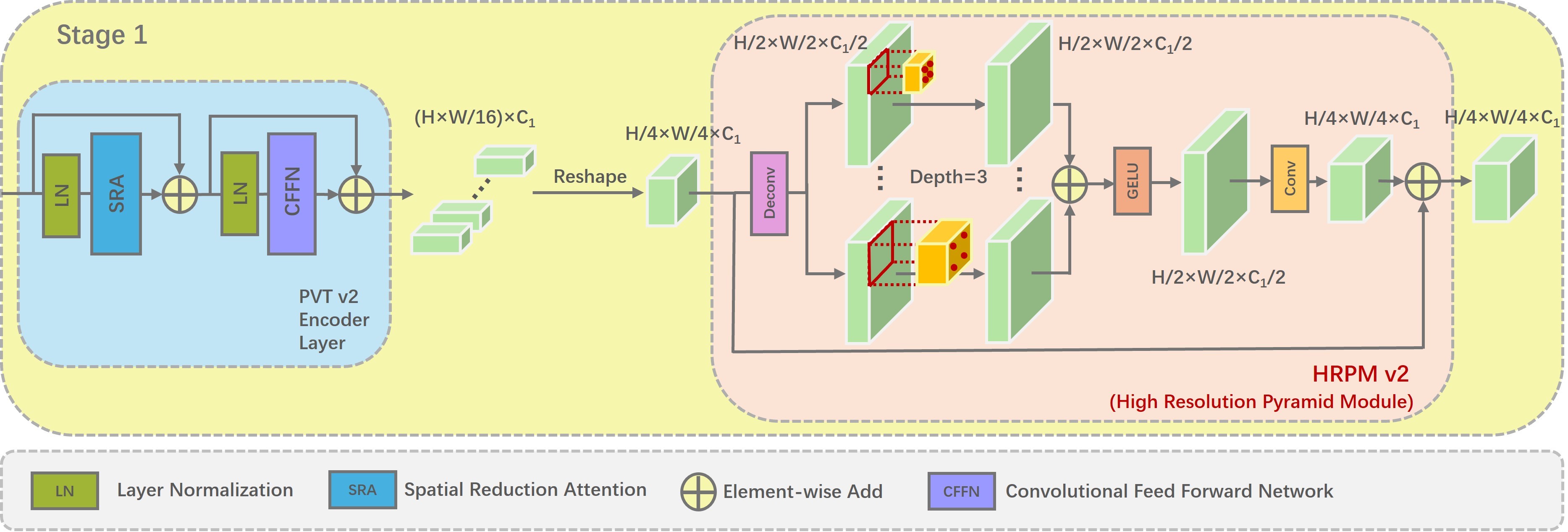}
\caption{Illustration of the structure of HRPM v2: In stage 2, after passing through all the PVT v2 encoder layers, the token sequence is first reshaped back into a feature map of size H/4×W/4×$C_1$, where $C_1$ is the channel number of the first stage. This feature map is then upsampled to H/2×W/2×$C_1$/2 using a deconvolutional layer. Next, it passes through a hierarchical hybrid-dilated convolutions structure with a depth of 3, and the high-resolution features are aggregated using an element-wise addition operation to model the high-resolution pyramid structure again. Finally, a strided convolutional layer downsamples the feature map back to its original size. Additionally, throughout the entire process, a residual branch is utilized to reuse the learned features from the previous stage.
}\label{fig: hrpmv2}
\end{figure*}
\subsection{High-Resolution Pyramid Module}

\subsubsection{HRPM v1}
HRPM comprises two submodules, HRPM v1 and HRPM v2. The structure of HRPM v1 is shown in Figure \ref{fig: hrpmv1}. Unlike PVTv2, which directly splits and flattens images into visual tokens using OPE with strides of 4 pixels in the stem-net, HRPM v1 employs progressive downsampling, i.e., using two convolutional layers each with a stride of 2 pixels, to extract representations of high-resolution images with finer granularity, thereby modeling the locality for high-resolution feature maps. HRPM utilizes hybrid-dilated convolutions (HDC) \cite{yu2015multi} with a hierarchical structure to capture spatial context information across multiple scales within varying receptive fields and model the scale-invariance, i.e.,
\begin{align}
f^{HRPM \, v1} & \triangleq HRPM\, v1(x) \notag \\
&= Conv(\sigma(\text{Cat}(HDC(x;k))))
\end{align}

where
\begin{equation}
HDC(x; k) = \left[ \phi^{d_1}(x); \ldots; \phi^{d_k}(x) \right].
\end{equation}
Here, $x \in \mathbb{R}^{H \times W \times 3}$ represents input image, $\sigma(\cdot)$ denotes the GELU \cite{hendrycks2016gaussian} activation function, $Cat(\cdot)$ signifies the concatenation operation, $Conv(\cdot)$ indicates the convolutional layer, which includes convolution, batch normalization, and ReLU \cite{agarap2018deep} activation function. $\phi^{d_1}(\cdot)$ denotes functions learned by $i^{th}$ dilated convolution and $HDC(\cdot)$ symbolizes the HDC structure. Specifically, in HRPM v1, the depth of HDC, $k$, is six layers, while in HRPM v2, it is three layers. We have empirically proven their effectiveness. After the HDC structure, the hierarchical features are concatenated along the channel dimension with GELU activation. Subsequently, passing through a convolutional layer, we obtain $f^{HRPM\text{ }v1} \in \mathbb{R}^{H/4 \times W/4 \times C_1}$. As the input to the first stage, $f^{HRPM\text{ }v1}$ needs to be reshaped into a 1D token sequence, which then enters the stacked SRA and CFFN to further encode the feature information, resulting in the first stage output $f_1$. The formulas are as follows:
\begin{equation}
s_1 = Img2Seq\left(f^{HRPM\text{ }v1}(x)\right),
\end{equation}
\begin{equation}
s_{1, j} = CFFN_{j-1}\left(SRA_{j-1}\left(s_{1, j-1}\right)\right) + SRA_{j-1}\left(s_{1, j-1}\right),
\end{equation}
\begin{equation}
f_1 = Seq2Img(s_{1, j}).
\end{equation}
Here, $Img2Seq(\cdot)$ flattens the feature map to a token sequence, $s_1 \in \mathbb{R}^{(H/4 \times W/4) \times C_1}$ represents the token sequence of first stage and $s_{1, j} \in \mathbb{R}^{(H/4 \times W/4) \times C_1}$ denotes the token sequence obtained after $s_1$ has passed through $j^{th}$ PVT v2 encoder layers. $SRA_{j-1}(\cdot)$ and $CFFN_{j-1}(\cdot)$ indicate the $(j-1)^{th}$ SRA and CFFN operations, respectively.
\begin{figure*}[t]
\centering
\includegraphics[width=\linewidth]{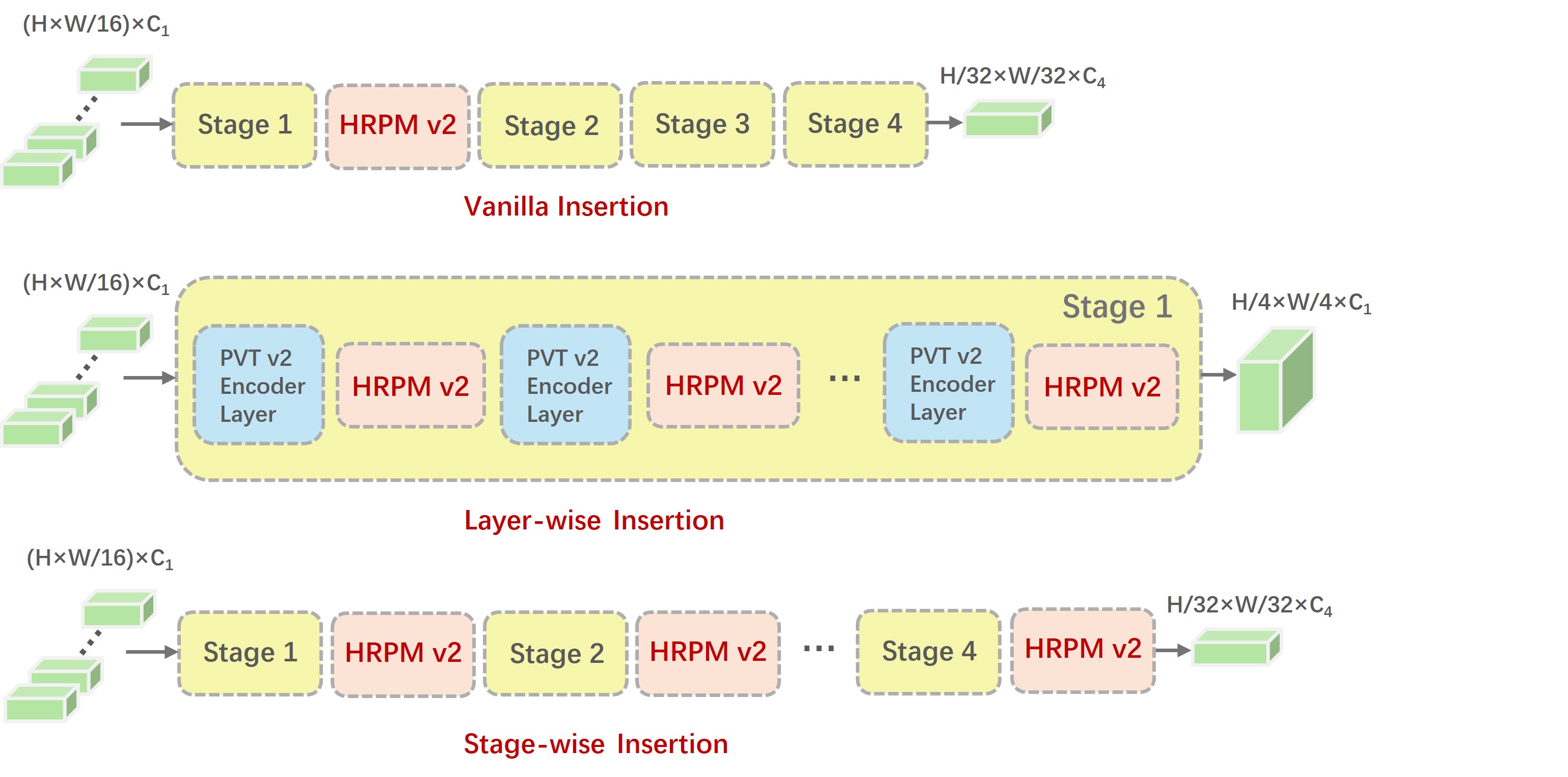}
\caption{Illustration of the three insertion methods for HRPM v2: Vanilla Insertion involves inserting only after the first stage. Layer-wise Insertion refers to inserting after each PVT v2 encoder layer within the first stage. Stage-wise Insertion involves inserting after each stage.
}\label{fig: two variants}
\end{figure*}
\subsubsection{HRPM v2}
Compared to directly employing progressive downsampling in HRPMv1, as shown in Figure \ref{fig: hrpmv2}, HRPMv2 first upsamples the feature map from a 1/4 scale to a 1/2 scale using a deconvolutional layer. It then further extracts high-resolution representations using non-downsampling HDC structure. It should be noted that after HDC, we use an element-wise addition operation to merge the hierarchical features, allowing the network to utilize both the high-resolution information from lower layers and the high semantic information from upper layers while maintaining the transmission of multi-scale features. Finally, progressive downsampling is applied, reducing the feature map from a 1/2 scale to a 1/8 scale. Additionally, a residual branch is utilized to reuse the learned features from the previous stage. The entire process can be described as follows:
\begin{align}
f^{HRPM \, v2} & \triangleq HRPM \, v2(f_1) \notag \\
&= Conv \left( \sigma \left( \sum_{i=1}^k HDC \left(\phi^{DEC}\left(f_1\right); k\right) \right) \right) + f_1
\end{align}
Here, $\phi^{DEC}(\cdot)$ denotes function learned by the deconvolutional layer. $f^{HRPM\text{ }v2} \in \mathbb{R}^{H/4 \times W/4 \times C_1}$ as the output feature map of HRPM v2, will serve as the input for the second stage for further processing. To this point, our vanilla HRPVT is completed. In addition to this, we have designed two variants. We will further elaborate on these in Section \ref{subsec: two} and validate their effectiveness in Section \ref{sec: ablation}.

\subsection{Two insertion strategies}
\label{subsec: two}
Since PVT v2 includes various models with different capacities, the performance improvement by using only vanilla HRPM is quite limited. Therefore, we designed two insertion strategies in accordance with the number and insertion position of HRPM v2, namely Layer-wise Insertion and Stage-wise Insertion. The structure of vanilla HRPVT and its two other insertion strategies are shown in Figure \ref{fig: two variants}. 

Layer-wise Insertion means inserting HRPM v2 after each PVT v2 encoder layer within the first stage only, aiming to extract richer high-resolution features compared to vanilla HRPVT without introducing excessive model complexity. The formula is as follows:
\begin{equation}
s_{1,j} = HRPM_{j-1}\left(Layer_{j-1}(s_{1,j-1})\right).
\end{equation}
Here, $Layer_{j-1}(\cdot)$ and $HRPM_{j-1}(\cdot)$ represents the $(j-1)^{th}$ PVT v2 encoder layer and HRPM v2, respectively. 

Stage-wise Insertion involves inserting HRPM v2 after each stage. This approach seeks to incorporate the inductive bias of CNNs into various stages while guiding the network's learning with higher-resolution representation information compared to the current stage. The formula is as follows:
\begin{equation}
s_{i} = HRPM_{i-1}\left(Stage_{i-1}(s_{1})\right).
\end{equation}
Here, $Stage_{i-1}(\cdot)$ represents the $(j-1)^{th}$ stage.

\section{Experiments}
\subsection{Datasets and evaluation metrics}
\subsubsection{MS COCO dataset}
The MS COCO dataset \cite{lin2014microsoft} comprises more than 200,000 images and 250,000 labeled person instances, each marked with 17 keypoints. Our model is trained using the MS COCO train2017 dataset, which includes 57,000 images and 150,000 person instances. We assess our model's performance on two subsets: the val2017 set, which has 5,000 images, and the test-dev2017 set, which includes 20,000 images. In the MS COCO dataset, the evaluation metrics employed are Average Precision (AP) and Average Recall (AR). These metrics are computed based on the Object Keypoint Similarity (OKS), which measures the alignment between the ground truth and the predicted keypoints. The formula is presented below: 
\begin{equation}
OKS = \frac{\sum_{i} \exp \left( -d_i^2 / 2s^2 k_i^2 \right) \delta (v_i > 0)}{\sum_{i} \delta (v_i > 0)}.
\end{equation}
In this metric, $d_i$ represents the Euclidean distance between a detected keypoint and its ground truth counterpart. The visibility of the ground truth keypoint is indicated by $v_i$, while s denotes the scale of the object, adjusting for size differences. Additionally, $k_i$ is a keypoint-specific constant that influences the rate of decay in the similarity measure based on the distance $d_i$. 
\begin{table*}[t]
\centering
\caption{Comparisons on the COCO val set. Pretrain = pretrain the backbone on the ImageNet classification task, except for the backbone of HRPVT-L, which is initialized with weights from mmpose. }
\label{tab: val}
\resizebox{\linewidth}{!}{%
    \begin{tabular}{llllllllllll}
    \hline
    Method           & Backbone         & Pretrain & Input size & \#Params & GFLOPs & AP   & AP$^{50}$ & AP$^{75}$ & AP$^{M}$           & AP$^{L}$  & AR   \\ \hline
    \multicolumn{12}{c}{2D Heatmap-based}                                                                                              \\ \hline
    SimpleBaseline \cite{xiao2018simple}   & ResNet-152       & Y        & 256 × 192  & 68.6M    & 15.7   & 72.0 & 89.3 & 79.8 & 68.7          & 78.9 & 77.8 \\
    HRNet-W48 \cite{sun2019deep}       & HRNet-W48        & Y        & 256 × 192  & 63.6M    & 14.6   & 75.1 & 90.6 & 82.2 & 71.5          & 81.8 & 80.4 \\
    TransPose-H-A4 \cite{yang2021transpose}  & HRNet-Small-W48  & Y        & 256 × 192  & 17.3M    & 17.5   & 75.3 & -    & -    & -             & -    & 80.3 \\ \hline
    TokenPose-T \cite{li2021tokenpose}     & pure Transformer & N        & 256 × 192  & 5.8M     & 1.3    & 65.6 & 86.4 & 73.0   & 63.1          & 71.5 & 72.1 \\
    TokenPose-B      & HRNet-W32-stage3 & Y        & 256 × 192  & 13.5M    & 5.7    & 74.7 & 89.8 & 81.4 & 71.3          & 81.4 & 80.0 \\
    TokenPose-L/D6   & HRNet-W48-stage3 & Y        & 256 × 192  & 20.8M    & 9.1    & 75.4 & 90.0 & 81.8 & 71.8          & 82.4 & 80.4 \\ \hline
    HRFormer-T \cite{yuan2021hrformer}      & HRFormer-T       & Y        & 256 × 192  & 2.5M     & 1.3    & 70.9 & 89.0 & 78.4 & 67.2          & 77.8 & 76.6 \\
    HRFormer-T       & HRFormer-T       & Y        & 384 × 288  & 2.5M     & 1.8    & 72.4 & 89.3 & 79.0 & 68.2          & 79.7 & 77.9 \\ \hline
    PoseTrans \cite{jiang2022posetrans}       & ResNet-101       & Y        & 256 × 192  & 53.0M    & 12.4   & 72.7 & 90.0 & 80.7 & 69.5          & 78.8 & 78.3 \\
    ViTPose \cite{xu2022vitpose}         & ViTPose-B        & Y        & 256 × 192  & 86.0M      & 17.1   & 75.8 & 90.7 & 83.2 & 68.7          & 78.4 & 81.1 \\
    PVT v2 \cite{wang2022pvt}          & PVT v2-B2        & Y        & 256 × 192  & 29.1M    & 5.1    & 73.7 & 90.5 & 81.2 & 70.0            & 80.6 & 79.1 \\ \hline
    MSPose-T \cite{yuan2024multi}        & TokenPose-T      & N        & 256 × 192  & 5.8M     & 1.3    & 67.1 & 87.3 & 75.3 & 64.4          & 73.0 & 73.4 \\
    MSPose-L         & TokenPose-L/D24  & Y        & 256 × 192  & 27.5M    & 11.0     & 76.0   & 90.5 & 82.7 & 72.8          & 82.9 & 81.2 \\ \hline
    LMFormer \cite{li2024lmformer}        & LMFormer-L       & N        & 256 × 192  & 4.1M     & 1.4    & 68.9 & 88.3 & 76.4 & -             & -    & 74.7 \\ \hline
    \multicolumn{12}{c}{SimCC-based}                                                                                                   \\ \hline
    PVT v2+ Simcc    & PVT v2-B2        & Y        & 256 × 192  & 25.0M      & 4.0      & 74.9 & 90.2 & 82.0   & 71.7          & 81.3 & 80.2 \\
    HRNet-W32+ Simcc & HRNet-W32        & Y        & 256 × 192  & 31.3M    & 7.8    & 75.3 & 90.2 & 81.9 & 71.9          & 81.7 & 80.8 \\
    HRPVT-S          & PVT v2-B0        & Y        & 256 × 192  & 4.8M     & 1.1    & 69.7 & 88.4 & 77.6 & 66.9          & 75.3 & 75.1   \\
    HRPVT-S          & PVT v2-B0        & Y        & 384 × 288  & 5.0M       & 2.7    & 73.3 & 89.3 & 80.3 & 70.1          & 79.6 & 78.7 \\
    HRPVT-L          & PVT v2-B2        & Y        & 256 × 192  & 25.1M    & 5.4    & 75.2 & 90.6 & 82.4 & \textbf{72.1} & 81.4 & 80.4 \\
    HRPVT-L          & PVT v2-B2        & Y        & 384 × 288  & 25.5M    & 12.5   & 76.3 & 90.6 & 83.2 & \textbf{72.9} & 82.8 & 81.5 \\ \hline
    \end{tabular}%
    }
\end{table*}

\begin{table*}[t]
\centering
\caption{Comparisons on the COCO test-dev set. \#Params and GFLOPs calculations exclude those related to human detection and keypoint grouping.}
\label{tab: test}
\resizebox{\linewidth}{!}{%
    \begin{tabular}{lllllllllll}
    \hline
    Method               & Backbone         & Pretrain & Input size & GFLOPs & AP   & AP$^{50}$ & AP$^{75}$ & AP$^{M}$           & AP$^{L}$  & AR   \\ \hline
    \multicolumn{11}{c}{2D Heatmap-based}                                                                                       \\ \hline
    CPN \cite{chen2018cascaded}                 & ResNet-Inception & Y        & 384×288    & 29.2   & 72.1 & 91.4 & 80.0   & 68.7          & 77.2 & 78.5 \\
    SimpleBaseline       & ResNet-152       & Y        & 384×288    & 35.6   & 73.7 & 91.9 & 81.1 & 70.3          & 80.0   & 79.0   \\
    TransPose-H-A6       & HRNet-Small-W48  & Y        & 256×192    & 21.8   & 75.0 & 92.2 & 82.3 & 71.3          & 81.1 & 80.1 \\
    TokenPose-L/D24      & HRNet-W48-stage3 & Y        & 256×192    & 11.0     & 75.1 & 92.1 & 82.5 & 71.7          & 81.1 & 80.2 \\
    HRNet-W48            & HRNet-W48        & Y        & 384×288    & 32.9   & 75.5 & 92.5 & 83.3 & 71.9          & 81.5 & 80.5 \\
    HRFormer-S           & HRFormer-S       & Y        & 384×288    & 6.2    & 74.5 & 92.3 & 82.1 & 70.7          & 80.6 & 79.8 \\ \hline
    \multicolumn{11}{c}{Regression-based}                                                                                       \\ \hline
    DeepPose \cite{toshev2014deeppose}            & ResNet-152       & Y        & 256×192    & 11.3   & 59.3 & 87.6 & 66.7 & 56.8          & 64.9 & -    \\
    PRTR \cite{li2021pose}                & HRNet-W32        & Y        & 384×288    & 21.6   & 71.7 & 90.6 & 79.6 & 67.6          & 78.4 & 78.8 \\
    RLE \cite{li2021human}                 & ResNet-152       & Y        & -          & -      & 75.1 & 91.8 & 82.8 & 72.0            & 80.2 & -    \\ \hline
    \multicolumn{11}{c}{SimCC-based}                                                                                            \\ \hline
    SimpleBaseline+Simcc & ResNet-50        & Y        & 384×288    & 20.2   & 72.7 & 91.2 & 80.1 & 69.2          & 79.0   & 78.0   \\
    HRNet-W48+Simcc      & HRNet-W48        & Y        & 256×192    & 14.6   & 75.4 & 92.4 & 82.7 & 71.9          & 81.3 & 80.5 \\
    HRPVT-S              & PVT v2-B0        & Y        & 384×288    & 2.7    & 72.5 & 91.2 & 80.2 & 69.8          & 77.9 & 77.9 \\
    HRPVT-L              & PVT v2-B2        & Y        & 384×288    & 12.5   & 75.5 & 92.4 & 83.3 & \textbf{72.5} & 80.9 & 80.7 \\ \hline
    \end{tabular}%
    }
\end{table*}
\begin{table*}[t]
\centering
\caption{Comparisons on the MPII test set. @0.5 means the threshold of the normalized distance is set 0.5.}
\label{tab: mpii}
\begin{tabular}{lllllllllll}
\hline
Method                & \#Params & GFLOPs & Hea           & Sho           & Elb           & Wri         & Hip           & Kne           & Ank           & PCKh@0.5 \\ \hline
SimpleBaseline-Res152 & 68.6M    & 28.7   & 96.8          & 95.4          & 89.3          & 83.9        & 87.9          & 84.6          & 81            & 88.9     \\
HRNet-W48             & 63.6M    & 35.4   & \textbf{97.2} & 95.7          & \textbf{90.6} & 85.6        & 89.1          & \textbf{86.9} & 82.3          & 90.1     \\
TranPose-H-A6         & 17.5M    & -      & -             & -             & -             & -           & -             & -             & -             & 90.3     \\
TokenPose-L/D24       & 28.1M    & -      & 97.1          & 95.9          & 90.4          & \textbf{86.0} & 89.3          & 87.1          & \textbf{82.5} & 90.2     \\
LMFormer-L            & 4.1M     & 1.9    & -             & -             & -             & -           & -             & -             & -             & 87.6     \\
HRPVT-S               & 4.9M     & 1.5    & 96.2          & 95.3          & 87.7          & 81.2        & 88.2          & 82.3          & 78.1          & 87.6     \\
HRPVT-L               & 25.3M    & 7.3    & 96.9          & \textbf{96.1} & 90.3          & 84.9        & \textbf{89.7} & 86.2          & 81.7          & 89.9     \\ \hline
\end{tabular}
\end{table*}
\subsubsection{MPII dataset}
The MPII Human Pose dataset \cite{andriluka20142d} is comprised of approximately 25,000 images featuring full-body pose annotations, captured across a diverse array of real-world activities. It includes around 40,000 subjects, with 12,000 designated for testing and the remainder used for training. The evaluation metric employed in the MPII Human Pose dataset is known as the Percentage of Correct Keypoints with head-based normalization (PCKh). This metric assesses the accuracy of predicted keypoints by determining whether the distance between a predicted keypoint and its corresponding ground truth keypoint falls within a predefined threshold. The approach to data augmentation and training strategies aligns with those employed for the MS COCO dataset, with the exception that images are cropped to a uniform size of 256 × 256 pixels to ensure consistent comparisons across different methods.

\subsection{Implementation details}
For the MS COCO Keypoint validation set, we first crop the input image based on the human detection box from SimpleBaseline \cite{xiao2018simple}, and resize the cropped boxes to either 256 × 192 or 384 × 288. Then, we perform data augmentation including horizontal flip, scaling (0.65, 1.35) and random rotation (-45°,+45°), while the MPII dataset uniformly resizes images to 256 × 256 for equitable comparisons with other approaches. Our models, with three different capacities, were trained and tested on two RTX2080Ti GPUs and one RTX3080Ti GPU based on the mmpose \cite{mmpose2020} codebase. In the case of 256 × 192, the backbones are initialized with PVT v2 official pre-trained weights, except for HRPVT-L, which is initialized with the weights from mmpose. The default training setting in mmpose is utilized for training the HRPVT models, i.e., we use Adam \cite{kingma2014adam} optimizer with a learning rate of 5e-4. After 170 epochs, the learning rate decreases by a factor of 10 during the subsequent 40 epochs and again in the final 10 epochs. The total training duration spans 210 epochs. In the case of 384 × 288, the backbones are initialized with the weights from the 256 × 192 configuration and fine-tuned for 100 epochs. It is important to highlight that the scaling factor $K$ for our S (small), M (medium), and L (large) models is set to 4.0, 4.0, and 6.0, respectively, which empirically performed better.
\subsection{Experimental results}
\subsubsection{Result on the MS COCO dataset}
The results of our method, along with those of other state-of-the-art methods on the MS COCO dataset, are reported in Table \ref{tab: val} and Table \ref{tab: test}. 

\paragraph{\textbf{Results on the validation set}} As shown in Table \ref{tab: val}, our method not only outperforms state-of-the-art 2D heatmap-based methods but also demonstrates a better trade-off between accuracy and complexity compared to SimCC-based methods. Specifically, our HRPVT-L achieves higher accuracy than HRNet-W48 while saving 60\% of the parameters and 63\% of the GFLOPs. Furthermore, when compared to recent state-of-the-art methods such as TransPose and TokenPose, HRPVT-L achieves comparable performance with fewer GFLOPs. In comparison to SimCC-based methods like those using HRNet-W32 as the backbone, HRPVT-L achieves a trade-off with only a 0.1 AP accuracy loss while saving 6.2M parameters and 2.4 GFLOPs. Additionally, when the resolution is increased to 384×288, HRPVT-L attains a higher accuracy of 76.3 AP, surpassing MSPose-L with a similar model capacity.

On the other hand, for the smaller model HRPVT-S, at a resolution of 256×192, it achieves 2.6 AP and 0.8 AP higher than MSPose-T and LMFormer-L, respectively. Although HRPVT-S does not outperform the state-of-the-art HRFormer-T at 256×192, it leads by 0.9 AP at 384×288, demonstrating its stronger discriminative ability with higher resolution inputs.
Notably, our method leads all models in Table \ref{tab: val} in the AP$^{M}$ metric, demonstrating HRPVT's superiority in medium and small-scale human pose estimation.

\paragraph{\textbf{Results on the test-dev set}} Table \ref{tab: test} presents a comparison of our method with other state-of-the-art methods. From the results, we can see that HRPVT continues to exhibit excellent performance on the more challenging test-dev set. Specifically, HRPVT-L achieves the same accuracy as HRNet-W48, but with only 38\% of its computational cost (GFLOPs). Additionally, although HRPVT-S has a 0.2 AP lower accuracy compared to the SimCC-based SimpleBaseline method, it achieves a 0.6 AP$^{M}$ improvement.
\subsubsection{Result on MPII dataset}
To evaluate the generalization capability of our method, we also conducted experiments on the MPII dataset, where all models were trained from scratch. As shown in Table \ref{tab: mpii}, despite having 0.8M more parameters, HRPVT-S achieves the same performance as LMFormer-L while reducing GFLOPS by 21\%. Although HRPVT-L does not surpass other state-of-the-art methods in terms of the overall PCKh@0.5 metric, it demonstrates better performance in detecting more challenging keypoints such as shoulders and hips.

\subsection{Ablation study}
\label{sec: ablation}
To verify the effectiveness of our HRPM, we conducted extensive experiments on the MS COCO Keypoint validation set, with a standardized input size of 256 × 192.
\subsubsection{HRPM gain breakdown}
We fine-tune HRPM v1 and HRPM v2 separately with a scaling factor $K$ of 2.0 to evaluate their independent contributions to our vanilla HRPVT. As shown in Table \ref{tab:depth}, there was no significant increase in accuracy with increasing depth until we set the depths of the two sub-modules to 6 and 3, respectively, resulting in a notable improvement in accuracy to 74.86 AP. However, further increases in depth led to a decrease in accuracy. We believe that incorporating multi-scale information appropriately during the high-resolution stage aids in network learning, while an excessive number of low-level features extracted from early stages are less beneficial. It's worth noting that the depth of HRPM v2 is set to half of HRPM v1 to maintain scale consistency and to prevent a significant increase in computational overhead without a corresponding improvement in accuracy. Additionally, we conducted a study on the width of HDC. As shown in Table \ref{tab:width}, the best accuracy was achieved when HRPM v1 and HRPM v2 were configured with 16 and 32 channels, respectively, they achieve relatively higher accuracy, whereas in other cases, there is a varying degree of decline. This indicates that the width of HDC is not necessarily better when larger. Finally, the combination of HRPM v1 and HRPM v2 resulted in the highest accuracy of 74.86 AP, underscoring the complementary nature of these two modules.
\subsubsection{HRPVT vs. two variants}
\begin{figure}[t]
\centering
\includegraphics[width=\linewidth]{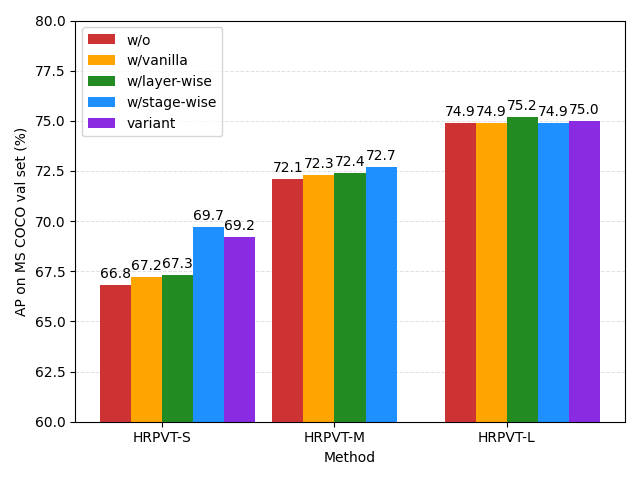}
\caption{Ablation study on three insertion methods and their variants. 'w/o' indicates the absence of HRPM, 'w/vanilla' refers to using Vanilla Insertion, 'w/layer-wise' and 'w/stage-wise' represent the use of the Layer-wise Insertion strategy and Stage-wise Insertion strategy respectively, and 'variant' denotes a variant of one of these two insertion strategies.}\label{fig: dilate}
\end{figure}
We conducted experiments on the proposed two insertion strategies. To further validate the effectiveness of HRPM, we designed two corresponding variants based on these strategies. Both variants adopt the same design principle: we removed HRPM v1 and replaced all deconvolutional layers and strided convolutional layers in HRPM v2 with point-wise convolutional layers \cite{lin2013network} to ensure consistency in the number of channels. All models were trained from scratch using the same training strategy.

As shown in Figure \ref{fig: dilate}, significant improvement (+2.9 AP) was observed when applying the Stage-wise Insertion strategy to the baseline PVT v2-B0 of HRPVT-S. Although the corresponding variant also performed well, it lacked HRPM and thus had a 0.5 AP gap compared to the original model. Meanwhile, we found that as the model capacity increased, the performance gains from the Stage-wise Insertion strategy diminished. This indicates that the strategy effectively utilizes higher-resolution representation information compared to the current stage to guide the learning of small networks. However, for a model with the capacity like HRPVT-L, richer high-resolution features are more necessary to improve the localization accuracy of medium and small-scale human keypoints, thereby enhancing overall performance. Consequently, a 0.3 AP improvement was observed with the Layer-wise Insertion strategy, while its variant only achieved a 0.1 AP improvement.
\begin{table}[t]
\centering
\caption{Ablation study on the HDC depth of our vanilla HRPVT, where "depth" represents the dilation rates ranging from 1 to n. For instance, a depth value of 3 indicates that HDC consists of three layers, with dilation rates set as [1, 2, 3] for each layer. }
\label{tab:depth}
\begin{tabular}{lll}
\hline
\multicolumn{2}{l}{HDC depth} & \multirow{2}{*}{AP} \\ \cline{1-2}
HRPM v1       & HRPM v2       &                     \\ \hline
$\times$      & $\times$      & 74.28               \\
2             & 2             & 74.58               \\
$\times$      & 3             & 74.59               \\
3             & 3             & 74.60               \\
4             & 3             & 74.61               \\
6             & $\times$      & 74.55               \\
6             & 3             & 74.86               \\
8             & 4             & 74.64               \\ \hline
\end{tabular}
\end{table}
\begin{table}[t]
\caption{Ablation study on the HDC width of our vanilla HRPVT, where "width" denotes the number of output channels for each layer in the HDC structure.}
\label{tab:width}
\centering
\begin{tabular}{llll}
\hline
\multicolumn{4}{l}{HDC width}     \\ \hline
HRPM v1 & AP    & HRPM v2 & AP    \\ \hline
16      & 74.86 & 16      & 74.76 \\
32      & 74.61 & 32      & 74.86 \\
64      & 74.78 & 64      & 74.70 \\ \hline
\end{tabular}
\end{table}

\section{Limitation and discussion}
\begin{figure*}[t]
\centering
\includegraphics[width=\linewidth]{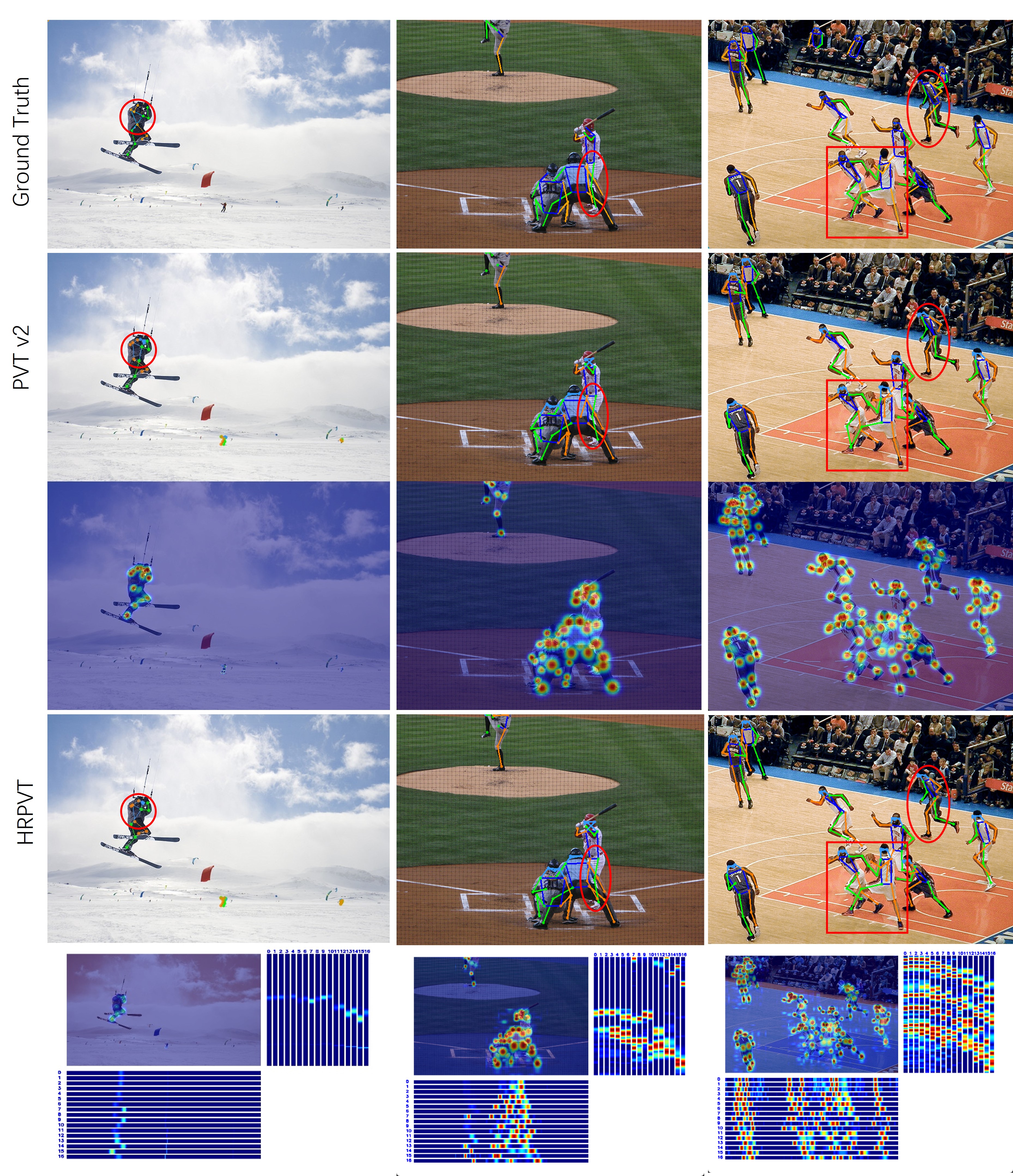}
\caption{Visualization and qualitative comparison of prediction results between HRPVT and PVT v2, from top to bottom: ground truth, PVT v2, and HRPVT, reveal that HRPVT exhibits better performance in estimating poses of medium and small sizes compared to PVT v2.
}\label{fig: compare}
\end{figure*}

The experimental results demonstrate the superiority of our proposed method in addressing medium and small-scale human pose estimation. However, when we further increased the model capacity, there was no significant performance improvement. We believe that our model is more adept at handling scenarios with limited computational resources and wide fields of view, such as outdoor sports capture. As shown in Figure \ref{fig: compare}, our method performs better in these scenarios.
\section{Conclusion}
In this paper, we propose HRPVT, a novel hybrid Vision Transformer architecture that uses a combination of PVT v2 and SimCC as its baseline. Building on this foundation, we designed the HRPM, which integrates the intrinsic inductive biases of CNNs into high-resolution feature maps to address the significant challenge of human pose estimation at medium and small scales. To accommodate models with varying parameter scales, we developed two distinct insertion strategies for HRPM, each tailored to enhance the model's ability to perceive medium and small-scale human poses from different perspectives. We conducted experiments on the MS COCO and MPII datasets, and the results demonstrated the effectiveness of HRPM and the two insertion strategies, showcasing the superiority of HRPVT in medium and small-scale human pose estimation. Future work involves extending HRPM to other hierarchical Vision Transformer architectures and enhancing its performance in models with larger capacity.

 \bibliographystyle{elsarticle-num} 
 \bibliography{elsarticle-template-num}

\end{document}